\pdfoutput=1

\documentclass[11pt]{article}
\usepackage{times}
\usepackage{latexsym}
\usepackage{url}

\usepackage{microtype}
\usepackage{amsmath}
\usepackage{graphicx}

\usepackage{subcaption}
\usepackage{url}
\usepackage{amsmath, amsthm, amssymb}

\usepackage{paralist}

\usepackage{enumitem}
\usepackage{xspace}

\usepackage{paralist}
\usepackage{booktabs} 

\usepackage{wasysym} 
\usepackage{array}

\usepackage{lingmacros}
\usepackage{extarrows} 

\usepackage{amsmath} 
\usepackage{multirow}
\usepackage[table,xcdraw]{xcolor}
\usepackage{ACL2024}
\usepackage{subcaption}
\usepackage{placeins}
\usepackage{times}
\usepackage{latexsym}
\usepackage{tabularx}
\usepackage{tabularray}
\usepackage{makecell}
\usepackage{arydshln} 
\usepackage{mdframed}
\usepackage{booktabs} 
\usepackage{array} 
\usepackage{ragged2e} 
\usepackage{adjustbox}
\usepackage{graphicx}
\usepackage{tcolorbox}
\usepackage{float}

\makeatletter
\newcommand{\medium}{%
    \@setfontsize\medium{9.0pt}{11.5pt}%
    \selectfont
}
\newmdenv[%
    backgroundcolor=gray!8,
    linecolor=black,
    outerlinewidth=0.5pt,
    roundcorner=1mm,
    skipabove=\topsep,
    skipbelow=\topsep,   font=\ttfamily\small,
]{examplebox}
\makeatother

\usepackage[T1]{fontenc}

\usepackage[utf8]{inputenc}

\usepackage{microtype}

\usepackage{inconsolata}

\title{Evaluating LLMs' Mathematical Reasoning in Financial Document Question Answering}

\author{
  Pragya Srivastava\textsuperscript{~\#}\thanks{\hspace{0.15cm}Work done during internship at Microsoft Research} ,\hspace{0.15cm}
  Manuj Malik\textsuperscript{\ddag},\hspace{0.15cm}
  Vivek Gupta\textsuperscript{§}\thanks{\hspace{0.15cm} Primary Mentor and Corresponding Author}~,\hspace{0.15cm}
  Tanuja Ganu\textsuperscript{\#},\hspace{0.15cm} 
  Dan Roth\textsuperscript{§} \\
  \textsuperscript{\#}Microsoft Research, 
  \textsuperscript{\ddag}Singapore Management University,
  \textsuperscript{§}University of Pennsylvania \\
  \texttt{\small \{t-pragyasri, taganu\}@microsoft.com}, \hspace{0.1cm}
  \texttt{\small manujm@smu.edu.sg}, \hspace{0.1cm} \texttt{\small \{gvivek,danroth\}@seas.upenn.edu}
}

\begin{document}
\maketitle
\begin{abstract}

Large Language Models (LLMs), excel in natural language understanding, but their capability for complex mathematical reasoning with a hybrid of structured tables and unstructured text remain uncertain. This study explores LLMs' mathematical reasoning on four financial tabular question-answering datasets: TATQA, FinQA, ConvFinQA, and Multihiertt. Through extensive experiments with various models and prompting techniques, we assess how LLMs adapt to complex tables and mathematical tasks. We focus on sensitivity to table complexity and performance variations with an increasing number of arithmetic reasoning steps. The results provide insights into LLMs' capabilities and limitations in handling complex mathematical scenarios for semi-structured tables. Ultimately, we introduce a novel prompting technique EEDP tailored to semi-structured documents, matching or outperforming baselines performance while providing a nuanced understanding of LLMs abilities.

\end{abstract}

\section{Introduction}

In the constantly evolving realm of artificial intelligence, Large Language Models (LLMs) have risen as cutting-edge tools for natural language understanding. They excel in a wide array of NLP tasks, including machine translation (MT), text summarization, question answering, and code generation. One specific area where LLMs' mathematical reasoning abilities come under scrutiny is the domain of numerical reasoning tasks. Past research has delved into the potential of language models for mathematical reasoning tasks, as seen in studies such as in \citet{amini-etal-2019-mathqa,upadhyay-chang-2017-annotating, patel-etal-2021-nlp, cobbe2021gsm8k}. These investigations provide a means to evaluate the performance of language models when it comes to solving mathematical problems, ranging from straightforward math word problems to more complex ones.

\begin{figure}[t]
\small
\centering
\begin{tcolorbox}[colback=white, arc=1mm, width=2.7in, colframe=gray, title=The Goldman Sachs Group Incorporation]
\textbf{Notes to Consolidated Financial Statements}\\
The table below presents a summary of Level 3 financial assets.\\[1em]
\begin{tabular}{|l|r|}
\hline
\rowcolor{gray!30}
\textbf{Financial Asset} & \textbf{Dec. 2017} \\
\hline
Cash Instruments          & \$15,395 \\
Derivatives               & \$3,802 \\
Other Financial Instruments & \$4 \\
\hline
\end{tabular}\\[1em]
\textbf{Q:} What was the total value of Level 3 financial assets for Goldman Sachs in December 2017?\\[1em]
\textbf{A:}  \colorbox{gray!30}{\$15,395 + \$3,802 + \$4 = \$19,201}

\end{tcolorbox}
\vspace{-1.0em}
\caption{\small An example of a semi-structured financial document question answering.}
\label{fig:example-fig}
\vspace{-1.5em}
\end{figure}

However, the problem becomes significantly more challenging when we encounter a hybrid of structured such as semi-structured tables and unstructured text, as shown in example in figure \ref{fig:example-fig}. Such tables are common in documents such as invoices, health records, and financial reports in corporate settings. In the financial domain, these tables present numerical data in a structured format, including income statements, balance sheets, cash flow statements, shareholder equity data, and annual reports. The majority of NLP models are primarily trained to handle raw unstructured textual data, which limits their ability to reason over semi-structured data, such as tables, or more intricate hybrids of tables and text, as seen in \citet{chen-etal-2020-hybridqa, aly-etal-2021-fact, 2019TabFactA, chen2021ottqa}. Tables, especially these with intricate relationships and dependencies, often necessitate multi-hop reasoning, connecting information across multiple steps, as shown in \citet{gupta-etal-2020-infotabs}.

NLP models may encounter difficulties in performing such multi-step reasoning, particularly when dealing with complex mathematical operations involving tables, as highlighted in \citet{li-etal-2022-finmath}.
Previous research such as \citet{chen-2023-large}, exemplified these issues and demonstrated LLM capacity to process and reason with semi-structured tables. However, these studies are somewhat constrained and don't explicitly explore the models' mathematical reasoning abilities. This is particularly evident in data/tasks that involve a substantial number of arithmetic reasoning steps, operate on extreme orders of magnitude, or deal with intricate tables where extracting the relevant information for a query becomes challenging. 

Moreover, when handling domain-specific documents, such as those in finance, a language model must not only have the necessary domain knowledge to craft the right approach for task-solving but also the capability to manipulate structured data, such as tables. Therefore, in this study, we aim to fill this gap by providing both qualitative and quantitative analyses of LLM's ability to reason on mathematical content on four finance datasets i.e. FinQA \cite{chen-etal-2021-finqa}, TATQA \cite{zhu-etal-2021-tat}, ConvFinQA \cite{chen-etal-2022-convfinqa}, and Multihiertt \cite{zhao-etal-2022-multihiertt}. These datasets feature questions demanding intricate numerical reasoning, combining semi-structured tables and text. Each dataset provides pre-annotated executable programs for precise answer retrieval. Our goal is to illustrate how model performance varies as the numerical complexity of the underlying data and the intricacy of the mathematical reasoning steps required to solve a query increase. Building upon these observations, we propose a novel approach termed (\textbf{E}licit $\longrightarrow$ \textbf{E}xtract $\longrightarrow$ \textbf{D}ecompose $\longrightarrow$ \textbf{P}redict) \emph{EEDP}, designed to deconstruct model responses into discrete components. This innovative method offers a deeper, more transparent insight into the numerical limitations of the model when tackling these tasks. Our contributions are as follows:

1.  We conduct a comprehensive robust evaluation of state-of-the-art Large Language Models (LLMs) for tabular (hybrid) question answering, with a specific focus on mathematical reasoning tasks, using public financial tabular datasets to establish a thorough performance benchmark.
    
2. Our analysis is thorough and multifaceted, encompassing both qualitative and quantitative aspects across several dimensions. We aim to provide nuanced insights into the strengths and limitations of LLMs in tabular (hybrid) question answering, especially in scenarios involving mathematical reasoning.

3. Building upon qualitative analysis, we introduce a novel and improved prompting method called \textbf{EEDP}. Our novel approach not only enhances our understanding of model weaknesses but also substantially enhances model performance compared to existing prompting methods across multiple models types.

Our metadata dataset and source code are available at \href{https://vgupta123.github.io/eedp}{https://vgupta123.github.io/eedp}.

\section{Metadata Annotations}

\label{sec:meta-data}
We annotated four tabular datasets: FinQA, TATQA, ConvFinQA, and Multihiertt with meta information related to a.) reasoning steps, b.) question category, c.) table length, d.) hierarchical complexity e.) missing information. \footnote{One author annotated the data, and the other checked for accuracy; we took stringent measures to minimize errors.} Below, we provide detailed information about these meta-data annotations:






\paragraph{1. Number of Reasoning Steps: } Including the count of arithmetic operations in questions is crucial. More operations reflect increased complexity in reasoning, and their interdependence offers insights into the models' proficiency. This annotation, applied across all four datasets, reveals their ability in handling intricate arithmetic tasks. Refer to Figure \ref{fig:steps} in Appendix \ref{subsec:datacoverage} for distribution of questions based on the number of reasoning steps involved.

\paragraph{2. Question Categorization: }  

In numerical reasoning, grasping the evolution from fundamental arithmetic to advanced operations is crucial, marking a shift in cognitive complexity. As questions advance, they typically involve more intricate combinations of operations and linguistic nuances. Our research identify both the capabilities and limitations of LLMs in understanding these concepts.

We establish 12 mathematical concept categories (Table \ref{table:question-categories}) with corresponding definitions, annotating each question. The dataset coverage across these categories is shown in Figure \ref{fig:qsplit-dataset} in Appendix \ref{subsec:datacoverage}. Notably, categories like \textsc{Division} and \textsc{Ratio} share similarities but differ in focus: \textsc{Division} involves the division operator, while \textsc{Ratio} encompasses ratios, fractions, and inverse problems. \textsc{Change in Ratio} questions add complexity with quantity changes requiring subtraction. Additionally, we omit \textsc{NEED-IN-DOMAIN-INFO} due to domain-specific knowledge focus and \textsc{TIME} questions due to limited sample size.

\begin{table}
\centering
\small
\begin{adjustbox}{max width=\textwidth}
\begin{tabular}{p{1.4cm}p{5.3cm}}
\toprule
\textbf{Concepts} & \textbf{Definition} \\
\midrule
\textsc{Sum}                       & Questions that require only the knowledge of addition. \\

\textsc{Difference}                & Questions that require only the knowledge of subtraction. \\

\textsc{Product}                   & Questions that require only the knowledge of multiplication. \\

\textsc{Division}                  & Questions that require only the knowledge of division. \\

\textsc{Ratio}                     & Questions that require knowing fractional forms, e.g., percentages, ratios. \\

\textsc{Change Ratio}              & Questions involving the difference between two fractional forms, e.g., percentage changes, difference in ratios. \\

\textsc{Range}                     & Questions requiring knowledge of the minimum and maximum of data observations. \\

\textsc{Compare}                   & Questions necessitating a comparison between mathematical quantities (e.g., greater than, less than). \\

\textsc{Average}                   & Questions needing knowledge of the average, used to calculate the central tendency of a group of data points. \\

\textsc{In-Domain-Info}            & Questions that require implicit knowledge to understand domain-specific mathematical formulations (e.g., return on investment (RoI), cost of goods sold (COGS), amortization rate, etc.). \\

\textsc{Time}                      & Questions explicitly involving mathematical operators for time-spans not in the table or context. \\

\textsc{Counting}                  & Questions requiring the counting of elements in a set or group of data points. \\
\bottomrule
\end{tabular}
\end{adjustbox}
\caption{\small Mathematical concept categories and definitions for studying LLM concept comprehension abilities.}
\label{table:question-categories}
\vspace{-1.0em}
\end{table}

\paragraph{3. Table Length:} Evaluating performance with larger supporting tables is crucial. Larger tables complicate multi-hop reasoning tasks by increasing the amount of information, making it harder to identify relevant evidence. We prioritize these annotations for datasets like FinQA and Multihiertt, where questions mainly use tables as supporting evidence.  Therefore, these annotations are confined to these datasets. In Multihiertt, when multiple tables support evidence, we consider the one with the highest row count i.e. maximum table length. The dataset distribution for Multihiertt and FinQA w.r.t table length (number of rows) is shown in Figure \ref{fig:Number of Rows} in Appendix \ref{subsec:datacoverage}.

\paragraph{4. Hierarchical Complexity: } In hierarchical tables, such as those in Multihiertt, evaluating model performance concerning the growing hierarchical complexity in cells with critical information becomes paramount. 
To tackle this, we annotate each example in Multihiertt with the hierarchy depth of cells containing relevant information.
For table with multiple relevant cells, we consider the cell with the highest hierarchical depth for our analysis. Our approach to estimating hierarchy depth is illustrated in Figure \ref{fig:hierarchy-depth}. Figure \ref{fig:hierarchical complexity}(a) in Appendix \ref{subsec:datacoverage} illustrate how we calculate hierarchical complexity for examples with multiple relevant rows at various hierarchical depths.

\paragraph{5. Missing Information: } Interpreting a table becomes challenging as the number of empty cells increases. Empty cells indicate missing or undefined information, leading to potential gaps in understanding. 

Assessing \textit{empty cell proportions} is crucial to quantify data ambiguity. More empty cells suggest higher uncertainty, which can hinder models' ability to derive meaningful insights and impact reasoning accuracy. In Multihiertt, where tables are hierarchical in nature and empty cells occur quite  frequently, we annotate examples with the empty cells percentage, contributing to our understanding of data ambiguity. For distribution of missing information (empty cells proportions) across datasets, refer to Figure \ref{fig:hierarchical complexity} (b) in Appendix \ref{subsec:datacoverage}.

\paragraph{Annotation Splits.} We prioritized complex numerical questions in our selection criteria, balancing this with resource constraints such as the LLM context length limits. We also took into account tables with deeper hierarchies in Multihiertt and multi-turn conversations in ConvFinQA. For TATQA, we utilized 45$\%$ of the development set by filtering out examples involving simple span selection. In the case of Multihiertt, we included 68$\%$ of the test set by excluding examples where the table length exceeds 40.  For FinQA and ConvFinQA, we employed the complete test and development sets, respectively.


\section{Experimental Results}

In this study, we choose to experiment with LLMs such as \texttt{GPT-3.5-Turbo}, \texttt{GPT-4}, \texttt{PaLM-540B}, \texttt{Mistral-7B-Instruct}\footnote{https://huggingface.co/mistralai/Mistral-7B-Instruct-v0.1}, \texttt{Llama-2-13B}\footnote{https://huggingface.co/meta-llama/Llama-2-13b-chat-hf} and \texttt{MAmmoTH-13B}\footnote{https://huggingface.co/TIGER-Lab/MAmmoTH-13B}. These LLMs are at the cutting edge for both open-source and closed models applications. Models like \texttt{MAmmoTH-13B} are specifically fine-tuned during pre-training to excel in mathematical reasoning tasks. For more detail about the the model choices refer to Appendix \ref{sec:model_selection}.

\paragraph{LLMs Prompting Methods:} For an instruction-tuned LLM, it's assumed that we give the model a \textit{task-specific} instruction $\mathcal{I}$ accompanied with a few (usually $k \in \{2, 4\}$) demonstrations $\mathcal{D_T}$ of a task T. We experiment with the following prompting techniques:

\vspace{0.5em}
(a.) \textbf{Direct: } In this setup, we explicitly instruct the models to abstain from providing explanations  and just return the final answer to the posed question. For this scenario, $\mathcal{D_T}$ contains $\{(p_i, q_i, a_i)\}_{i=1}^{k} $ where p is the premise (table-text), q is the question, and a is the ground-truth answer.

\vspace{0.5em}
(b.) \textbf{CoT: } Moving forward, we experiment with the \textit{chain-of-thoughts} prompting strategy where  we instruct the model to output the explanation to the answer derived by it.  Here, our $\mathcal{D_T}$ contains $\{(p_i, q_i, e_i)\}_{i=1}^{k} $ where p is the premise which includes the table and the associated text, q is the question and  e is the explanation of the answer. 

 \vspace{0.5em}
(c.) \textbf{PoT: } In this case, the expected response is a code derivation of the answer.  Here, $\mathcal{D_T}$ contains $\{(p_i, q_i, c_i)\}_{i=1}^{k} $ where p is the premise which includes the table and the associated text, q is the question and  c is the code-derivation of the answer.
    
\vspace{0.5em}
(d.) \textbf{Decomposers:} \cite{ye2023large} proposed to address the challenge of handling large tables by decomposing them into more manageable subtables. Similarly, complex questions are handled by breaking them down into simpler subquestions. Decomposition proves effective with SQL tables, facilitating the removal of distracting details while retaining all supporting evidence. Questions are first parsed to break them down into simpler, more manageable subquestions. The model then addresses each subquestion independently before composing the answers to arrive at the final solution. In this case, our demonstration set $\mathcal{D_T}$ contains $\{(p_i^{'}, \langle q_1, q_2,..., q_n \rangle, a_i)\}_{i=1}^{k} $ where $p^{'}$ is the premise obtained by the irrelevant information removal to the question from the original premise p and $ \langle q_1, q_2,..., q_n \rangle$ are the subquestions whose answers lead to the final answer. 

\paragraph{EEDP Prompting Strategy: } We propose a \textit{novel prompting strategy}:  \textbf{E}licit $\longrightarrow$ \textbf{E}xtract $\longrightarrow$ \textbf{D}ecompose $\longrightarrow$ \textbf{P}redict.  Figure \ref{fig:methodfigure} show an illustration of our EEDP approach. Below are the detail of each EEDP step:

\begin{enumerate}
 \setlength{\itemsep}{0.5pt}
\item \textbf{Elicit:} We prompt the model explicitly to first \textit{elicit} relevant domain knowledge for answering a  given query.
\item \textbf{Extract:} Conditioned on the table, question and the elicited domain knowledge, the model extracts supporting evidences to answer a given question.
\item \textbf{Decompose:} We instruct the LLM to break a complex mathematical reasoning task into multiple atomic operations and compose the operations to arrive at the final answer.
\item \textbf{Predict:} The model finally returns the derived answer in the above steps.\end{enumerate}

Figure \ref{eedp-prompt-template} shows a example for EEDP strategy with one shot.

\begin{table*}[!ht]
    \centering
    \small
    \begin{tabular}{l@{\hspace{12mm}}lccccc}
\toprule
        \textbf{Dataset} & \textbf{Model} & \textbf{Direct} & \textbf{CoT} &    \textbf{PoT} & \textbf{EEDP} & \textbf{Decomposers} \\
\midrule
       &   \texttt{GPT-4} & 55.81  & 86.91 & \textbf{89.99} & 88.67 & 47.46 \\
       & \texttt{GPT-3.5-Turbo} & 31.38  & 77.57 & \textbf{82.11} & 79.73 & 28.53  \\
\textbf{TATQA}      & \texttt{PaLM 2-540B} & 44.66  & 62.93 & 61.60 & \textbf{81.51} & 57.94  \\
     &\texttt{Llama 2-13B} & 3.36   & 35.95 & 34.16 & \textbf{40.95} & 25.93 \\
      &  \texttt{MAmmoTH-13B} & 19.11  & \textbf{56.25} & 10.02 & 4.37  & 22.89 \\
      &  \texttt{Mistral-7B} & 10.92  & \textbf{59.14} & 16.53 & 56.06 & 7.24  \\     \midrule
      & \texttt{GPT-4} &  65.12  & 72.38 & 75.26 & \textbf{76.05} & 44.93\\
      & \texttt{GPT-3.5-Turbo} & 40.47  & 59.18 & \textbf{68.97} & 61.88 & 32.33  \\
\textbf{FinQA}       & \texttt{PaLM 2-540B} & 30.33  & 34.79 & 30.41 & \textbf{61.95} & 46.38  \\
     & \texttt{Llama 2-13B} & 1.80   & 25.34 & 12.97 & \textbf{30.47} & 11.91 \\
     & \texttt{MAmmoTH-13B} & 22.83  & \textbf{35.32} & 15.86 & 35.05 & 17.65 \\
      & \texttt{Mistral-7B} & 26.11  & 34.23 & 10.56 & \textbf{34.86} & 12.34 \\
      \midrule
 & \texttt{GPT-4} & 63.10  & 71.19 & \textbf{78.81} & 77.91 & 18.76 \\
  & \texttt{GPT-3.5-Turbo} & 37.62  & 48.33 & 61.19 & \textbf{61.75} & 10.50 \\
\textbf{ConvFinQA}   & \texttt{PaLM 2-540B} & 20.19  & 38.00 & 40.14 & \textbf{63.42} & 22.32   \\
  & \texttt{Llama 2-13B} & 3.80   & 29.45 & 29.92 & \textbf{39.42} & 10.35 \\
  & \texttt{MAmmoTH-13B} & 21.61  & \textbf{46.08} & 8.78  & 32.77 & 7.83  \\
  & \texttt{Mistral-7B} & 12.35  & \textbf{48.45} & 14.48 &  36.57 & 11.16 \\
  \midrule
 & \texttt{GPT-4} & 41.35  & 55.13 & 67.23 & \textbf{70.32} & 36.86  \\
 & \texttt{GPT-3.5-Turbo} & 25.88  & 42.33 & \textbf{52.18} & 49.65 & 20.61  \\
\textbf{Multihiertt} & \texttt{PaLM 2-540B} & 14.20  & 20.67 & 36.52 & \textbf{37.97} & 20.19  \\
 & \texttt{Llama 2-13B} & 1.54   & \textbf{30.66} & 18.12 & 24.15 & 16.86 \\
 & \texttt{MAmmoTH-13B} & 10.12  & \textbf{18.56} & 6.57  & 18.36 & 11.87 \\
 & \texttt{Mistral-7B} & 14.909 & \textbf{22.92} & 14.94 & 10.97 & 11.63 \\
        \bottomrule
    \end{tabular}
    \vspace{-0.5em}
    \caption{\small Comparison of performance of different models tested against a variety of prompting strategies}
    \label{accuracy-table}
    \vspace{-1.0em}
\end{table*}

\paragraph{Results and Analysis.}

Table \ref{accuracy-table} shows a comparison in performance between different prompting strategies. Despite being a single prompt, EEDP demonstrates comparable or superior performance compared to PoT. Notably, we outperform PoT significantly for \texttt{PaLM-2-540B} and \texttt{LLAMA-2-13B} across most datasets. Moreover, while PoT relies on external tools for executing mathematical programs/code to obtain answers, EEDP exclusively utilizes LLM for all tasks, including evidence extraction, operation identification, and execution, ensuring precision throughout the process.


As shown in Table \ref{accuracy-table}, the Decomposers prompting strategy exhibits a much poorer performance compared to other strategies. The reason behind this was statistically found to be the inaccurate formation of subtables from the main table, leading to information loss as described in the previous paragraph. The performance of EEDP either surpasses or matches very closely with that of PoT. The number of shots was adjusted depending on the context length of the model. 

We can see that \texttt{MAmmoTH-13B} model, which is fine-tuned on the \texttt{MathInstruct} dataset \cite{yue2024mammoth} containing Instruction-Response pairs where the responses are a hybrid of CoT and PoT rationales, fails to perform well with the EEDP methodology. We argue that this is due to two potential reasons: (a.) Reduction of the number of shots to adjust the context length as the EEDP response is longer than that of the other methods, and (b.) Finetuning may contribute to suboptimal performance due to its alignment with a particular style and format of responses, potentially limiting the model's adaptability and generalization to other diverse contexts.

\paragraph{EEDP's Computational Efficiency} EEDP functions as a unified single-prompt method, minimizing computational complexity. Unlike methods like PoT, which rely on external tools, EEDP operates independently. When assessing computational cost, we consider API calls and token generation. Since EEDP uses a single-step prompting approach, only one API call is needed per query, making its computational cost comparable to methods like CoT. For inference with open-source models, we used hardware with an A40 40GB GPU. Processing one dataset per model using the vLLM library took approximately 10 hours.


\section{Where do LLMs fail?}

Through manual inspection, we rigorously evaluate the EEDP responses against the meta-annotations from section \ref{sec:meta-data} as ground-truth benchmarks for extraction and model reasoning accuracy. The reasoning programs represent sequences of arithmetic operations necessary to derive the final answer, utilizing values extracted from supporting evidence as operands. To assess calculation accuracy, we manually identify the model's instantiation and precision errors. Our EEDP prompt ensures that the model predominantly outputs responses in the expected format, with exceptions being rare.
However, since we manually analyze all outputs, we do not penalize the model for format deviations but rather for incorrect outputs. Penalties are applied only when the model makes errors in extraction, reasoning, and/or calculation. Below, we categorize the EEDP response errors in detail based on their origins:

\begin{table}[!htbp]
\centering
\small
\begin{tabular}{l@{\hspace{2mm}}llr}
\toprule
\textbf{Dataset} & \textbf{Error} & \textbf{Type} & \textbf{Per.(\%)} \\
\midrule
\multirow{5}{*}{\textbf{FinQA}} & \multirow{2}{*}{{Extraction}} & E1 & 10.38 \\
 &  & E2 & 25 \\
 & {{Reasoning}} & R1 & 25 \\
 &  & R2 & 15.57 \\
 & Calculation &  C1/C2 &  24.06 \\
\midrule
\multirow{5}{*}{\textbf{ConvFinQA}} & \multirow{2}{*}{{Extraction}} & E1 & 8.45 \\
 &  & E2 & 14.08 \\
 & \multirow{2}{*}{{Reasoning}} & R1 & 19.72 \\
 &  & R2 & 36.62 \\
 & Calculation &  C1/C2 & 21.13 \\
\midrule
\multirow{5}{*}{\textbf{TATQA}} & \multirow{2}{*}{{Extraction}} & E1 & 13.79 \\
 &  & E2 & 31.03 \\
 & \multirow{2}{*}{{Reasoning}} & R1 & 22.41 \\
 & & R2 & 5.17 \\
 & Calculation &  C1/C2 & 27.59 \\
\midrule
\multirow{5}{*}{\textbf{Multihiertt}} & \multirow{2}{*}{{Extraction}} & E1 & 20.5 \\
 & & E2 & 31.5 \\
 & \multirow{2}{*}{{Reasoning}} & R1 & 15.5 \\
 & & R2 & 12 \\
 & Calculation & C1/C2 & 20.5 \\
\bottomrule
\end{tabular}
\vspace{-0.5em}
\caption{\small Error Analysis on Various Datasets. In this table, Extraction.E1: Missing Evidences, Extraction.E2: Wrong Evidences, Reasoning.R1: Insufficient Domain Knowledge, Reasoning.R2: Question Misinterpretation, Calculation: Instantiation (C1) and Precision errors (C2)}
\label{tab:error_analysis}
\vspace{-1.0em}
\end{table}

\paragraph{1. Incorrect Extraction:} This category encompasses errors where the model faces difficulties in accurately identifying and extracting the pertinent information necessary for effective problem-solving. These errors point to challenges in retrieving precise information. These errors can further be subdivided into two categories
\begin{itemize}
    \item \textbf{Missing/Incomplete Evidences (E1): } The model fails to extract all the necessary evidences which serve as ingredients to derive the final answer.
    \item \textbf{Wrong Evidences (E2): } The model extracts wrong values for variables as supporting evidences from the premise.
\end{itemize}

\paragraph{2. Incorrect Reasoning:} Errors in reasoning occur when the model struggles to formulate an appropriate and contextually relevant approach to tackle a given problem. Possible reasons include a lack of domain knowledge or an inaccurate interpretation of the posed question. Consequently, this error type can arise from two sources.
\begin{itemize}
    \item \textbf{Deficit in Domain Knowledge (R1): } These errors occur when the model attempts to derive an answer to the posed question using a wrong formula for \textit{domain-specific} measures, for eg. COGS, ROI etc.
    \item \textbf{Question Misinterpretation (R2): } These errors occur when the model interprets the question differently and provides responses that are not aligned with the intended query. Overall, the model's outputs lack grounding in the original question posed to it.
\end{itemize}


\paragraph{3. Incorrect Calculation: }This variety of errors include those where the model commits mistakes due to calculation mistakes. This can be of two types as described below.
\begin{itemize}
    \item \textbf{Incorrect Instantiation (C1): } These include cases if the model extracts the right evidences, uses the right derivation formula but instantiates the variables incorrectly with the values resulting in an incorrect answer.
    \item \textbf{Precision Error (C2): } Language models employ mathematical algorithms for arithmetic operations, but their results may not always be perfectly accurate due to insufficient data pattern coverage or introduced biases during training. Consequently, they can sometimes generate outputs with slight inaccuracies or deviations from correct results. We show a detailed analysis in \ref{subsec:oom}.
\end{itemize}



\textit{Analysis:} The above categorization provides a nuanced understanding of the diverse challenges and shortcomings exhibited in different facets of mathematical reasoning. We observe that in a lot of cases, the error propagates because of a deficiency in domain knowledge. It is critical for both evidence extraction and reasoning. Despite possessing general domain knowledge owing to the massive amount of data these models have been pre-trained upon, these models may require explicit prompts to elicit the specific domain knowledge needed for a particular question. Furthermore, errors can arise due to the models' limited proficiency in multi-step reasoning, especially in tackling questions involving multiple arithmetic operations in a sequence.

We give a quantitative measure of each type of errors for each of the 4 datasets we consider for our study in Table \ref{tab:error_analysis}. We also provide examples corresponding to each error category in figures \ref{Missing Evidences}, \ref{Wrong Evidences}, \ref{DK-Deficit}, \ref{Incorrect-Interpret}, \ref{Incorrect-Instantiation} and \ref{Incorrect-Calculation}. Statistically, we find that reasoning errors contribute a significant chunk to the total number of errors. In case of complex hierarchical tables like that in Multihiertt, the model is found to struggle with extracting the right supporting evidences from the premise for a given question. Calculation errors can be taken care of if a third-party calculation tool (an external agent) is chained to the language model. 

\begin{figure*}[t]
\centering
 \includegraphics[width=0.89\textwidth]{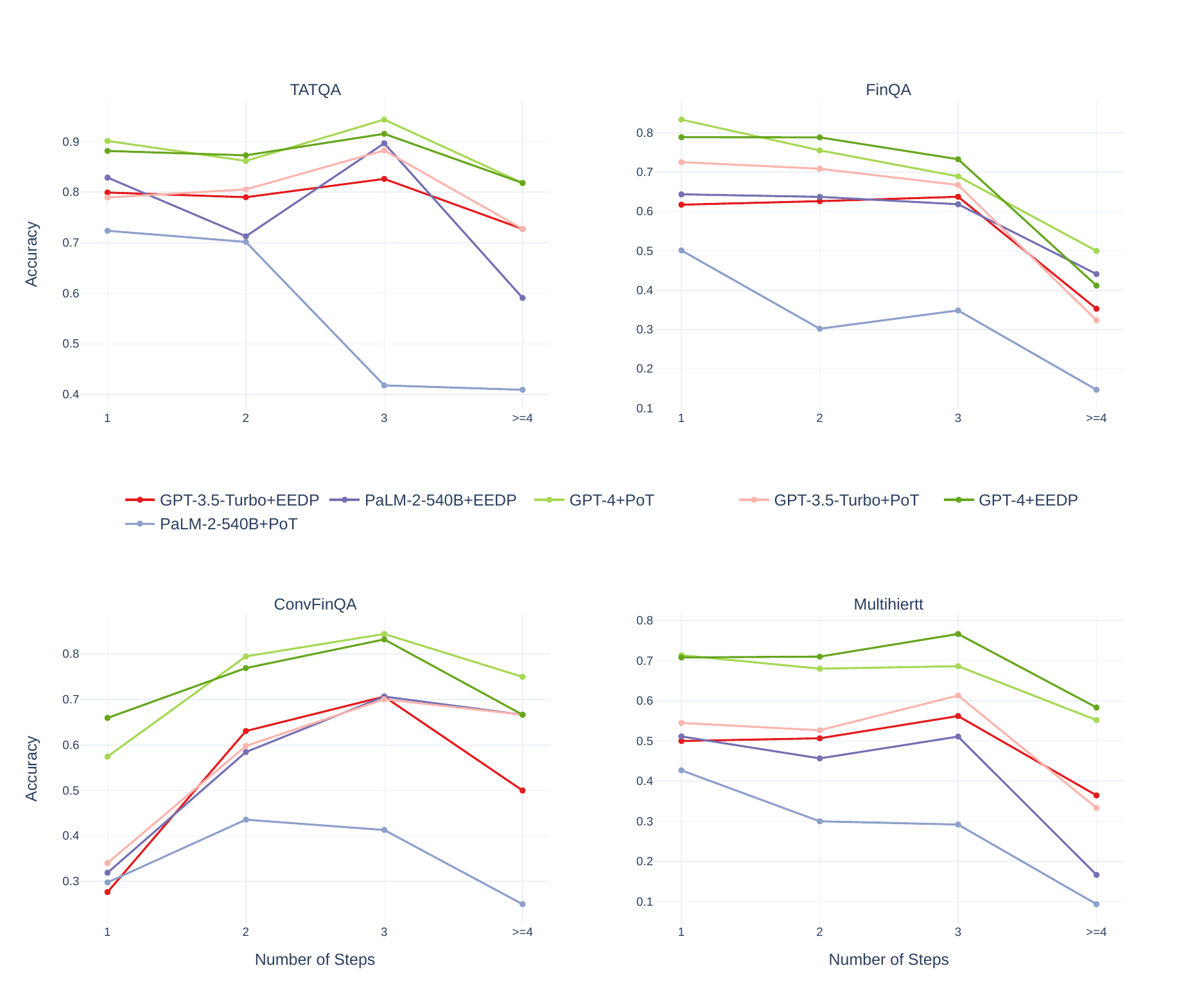}
    \vspace{-1.0em}
    \caption{\small A comparison showcasing the performance trends across various datasets with the increasing number of reasoning steps. The analysis contrasts the effectiveness of EEDP (our method) against  PoT in addressing complex reasoning.}
    \label{fig:Accuracy vs Steps}
    \vspace{-1.25em}
\end{figure*}

\begin{figure*}[t]
\centering
\includegraphics[width=0.98\textwidth]{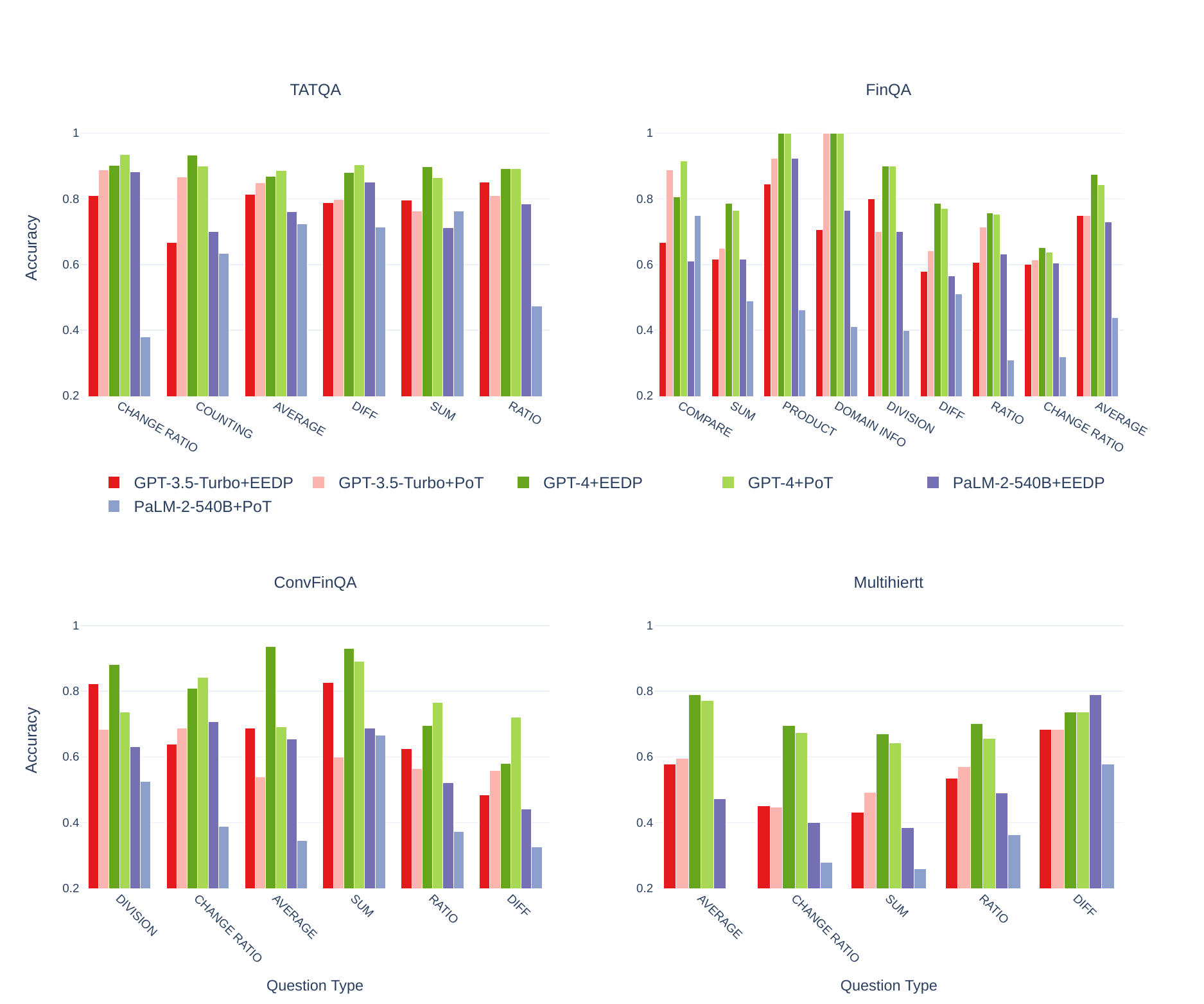}
    \vspace{-1.0em}
    \caption{\small A comparison showcasing the performance trends observed in various datasets across different question types. The analysis contrasts the effectiveness of EEDP (our method) against Few-Shot PoT (PoT). Best viewed in color.}
    \label{fig:Accuracy vs QSplit}
    \vspace{-1.25em}
\end{figure*}
 
\label{sec:datasets and exp}
\section{Analysis on Reasoning Annotations}
\label{sec:analysis}

We analyse model performance on the basis of fine-grained annotations as described in the section \ref{sec:datasets and exp}.

\paragraph{1. Performance vs Number of Reasoning Steps.} 


We investigate model performance with increasing mathematical reasoning steps, as shown in Figure \ref{fig:Accuracy vs Steps}. This analysis provides insights into models' ability to handle varying task complexities. As expected, performance decreases with more reasoning steps, indicating LLMs' challenges in retrieving information and reasoning as complexity grows. 

Anomalies are observed in ConvFinQA, where accuracy improves after greater than or equal to two reasoning steps, potentially due to questions referring to answers of prior conversation turns. Anomalies like these warrant further investigation beyond this study's scope.

\paragraph{2. Performance across Question Types.} 
We analyze the performance trends across different question categories, as defined in Table \ref{table:question-categories}, to assess the models' understanding of various mathematical and financial concepts. Figure \ref{fig:Accuracy vs QSplit} shows that EEDP consistently performs better than or as well as PoT across all datasets. The improvement is particularly pronounced for \texttt{PaLM-2-540B} in all question categories.
\paragraph{3. Performance across Arithmetic Operations.}
Figure \ref{fig:enter-oom} in Appendix \ref{subsec:oom} shows that for relatively simpler arithmetic operations like addition and subtraction, the effect of order of magnitude of the operands is less profound as compared to harder arithmetic operations such as multiplication and division. We observe the trend in the performance accuracy with the growing and diminishing orders of magnitude. We also observe bigger and more capable models such as \texttt{GPT-4}, \texttt{GPT-3.5-TURBO} and \texttt{PaLM 2-540B} perform much better on simpler addition, subtraction task in comparison to multiplication, division task. For more details on refer to the Appendix \ref{subsec:oom}.

\section{Other Related Works}
\label{sec:comparison}

\subsection{LLMs on Mathematical Reasoning} 
Pre-trained Language Models (PLMs) excel in NLP tasks \cite{devlin-etal-2019-bert, zhuang-etal-2021-robustly} by leveraging extensive textual corpora to acquire world knowledge \cite{10.5555/3524938.3525306}. Expanding PLMs for math-related tasks has been challenging due to their non-specific training. Recent attempts include MWP-BERT and Minerva \cite{liang-etal-2022-mwp, NEURIPS2022_18abbeef}, but curating high-quality math data remains difficult. To bridge the gap, researchers fine-tune PLMs for specific math tasks. Notable works, like Bhaskara, Self-sampler, Aristo, FinQANet, TAGOP, MT2Net, and others \cite{mishra-etal-2022-lila, ni2022learning, clark2021f, chen-etal-2021-finqa, zhu-etal-2021-tat, zhao-etal-2022-multihiertt, cao-xiao-2022-augmented, welleck2022naturalprover}, employ PLMs such as GPT-Neo and RoBERTa for math problem-solving.

\subsection{Tabular Question Answering}

Handling diverse input formats in question answering, including structured tables and visual data, poses challenges for language models. HybridQA \cite{chen-etal-2020-hybridqa} introduces questions requiring reasoning over tables and text. MultimodalQA \cite{talmor2021multimodalqa} adds visual inference. Our focus is on multi-hop question answering over tables and text. TAPAS \cite{herzig-etal-2020-tapas} tackles table-based questions without logical forms, while Tapex \cite{liu2022tapex} empowers generative models with table reasoning. 

Models like FinQANet, TagOP, and MT2Net \cite{chen-etal-2021-finqa, zhu-etal-2021-tat, zhao-etal-2022-multihiertt} employ a retriever module to extract supporting facts from input financial reports, followed by a reasoning module to derive the final answer. Retrieving relevant evidence and conducting reasoning both demand domain-specific knowledge, such as understanding financial transactions, identifying revenue trends, and interpreting complex investment statements. Thus, reliance on an external knowledge base becomes crucial for addressing the challenges of domain-specific multi-hop question answering.


\subsection{Prompt Engineering}

In-context Learning (ICL) equips Language Models (LLMs) with task examples and queries, enabling them to perform target tasks without updating model parameters \citep{NEURIPS2020_1457c0d6, openai2023gpt4}. They excel in mathematical reasoning with few-shot prompts but struggle with more complex tasks. Methods like chain-of-thoughts (CoT) \citep{NEURIPS2022_9d560961} have improved LLM performance by guiding them through intermediate reasoning steps. Enhancing multi-step reasoning involves two main approaches: improving in-context examples and obtaining better reasoning steps. Some focus on stable in-context example selection \citep{fu2023complexitybased, rubin-etal-2022-learning, lu2023dynamic}. Others adopt a modular approach, using off-the-shelf tools \citep{schick2023toolformer}, \textit{program of thoughts} (PoT) \citep{chen2022program}, or decomposition strategies \citep{zhou2023leasttomost, dua-etal-2022-successive, khot2023decomposed}.

Our task requires complex multi-step reasoning across diverse information sources. LLMs, as demonstrated by \citep{chen-2023-large}, can reason over structured tables without explicit encoding. They also serve as versatile decomposers, breaking down extensive evidence and complex questions \citep{ye2023large}.

\section{Key Takeaways}

\paragraph{Our Contributions.}  
Our study aimed to delve into the mathematical reasoning abilities of LLMs within the context of financial documents where models are tasked with complex hybrid (table-text) question answering. This presents a significant challenge, requiring models not only to provide accurate numerical analysis but also to retrieve right supporting evidence tailored to specific question requirement. Moreover, it necessitates the extraction of necessary knowledge from the model's pre-trained parameters to address queries.


Firstly, we meticulously annotate popular financial datasets, such as FinQA, ConvFinQA, TATQA, and Multihiertt, with detailed meta-information. This includes specifying mathematical operations, types of reasoning involved, table dimensions, question types, and the depth of table hierarchy. Furthermore, we conduct a manual error analysis to quantify error types across multiple LLMs. These detailed annotations are invaluable for analyzing various dimensions where LLM models encounter challenges. This, in turn, aids in the development of better prompting techniques such as EEDP, aimed at enhancing LLMs' mathematical reasoning abilities. The resulting improvement in performance with EEDP across multiples datasets serves as compelling evidence of the effectiveness of this approach. 

\paragraph{What did we learn?}


Our analysis revealed that LLMs can accurately handle addition and subtraction tasks e.g. modle perform fairly when calculating total expenses or profits, but struggle with multiplication and division e.g. model performs poorly with questions requiring reasoning operations involving proportions, ratios, percentages, and division.
Moreover, as the complexity of the data increases either through a higher absolute order scale or more decimal numbers, model performance degrades. Model performance also degrades with increasing number of reasoning steps and lengthy complex hierarchical table structures. e.g. in complex datasets with hierarchical structures such as Multiheirtt, TATQA, incorrect extraction leads to modeling errors. Similarly, on queries involving multiple conversational turns, such as those in ConvFinQA, model perform poorly due to reasoning failures, like misinterpreting multiple queries longer context. Across all models, incorrect reasoning and incorrect extraction consistently emerge as common sources of errors. For smaller models, even straightforward calculations, result in errors due to imprecise calculations.

\paragraph{EEDP vs other methods}
    \paragraph{(a.) EEDP vs PoT:} PoT enhances LLM inference with the use of variable names for the supporting values extracted from the premise and prompts the LLM to express their thought process in the form of programs. The model output is a program which is executed externally to derive the final answer. EEDP proposes to decompose a complex reasoning task into simple atomic steps whose solutions can be composed to give the final answer. In PoT, they don't make the language model do the computation while in our case the language model not only outputs the reason but also computes the final answer. This distinction implies that PoT may have an inherent advantage over EEDP. 
    \paragraph{(b.) EEDP vs Decomposers: } The prompting strategy proposed by \cite{ye2023large} was originally designed for querying SQL tables, they use LLMs to break down evidence and questions for SQL interpreters. In contrast, our approach addresses more complex scenarios involving both tables and text, requiring advanced reasoning skills beyond the capability of standard SQL interpreters. Pruning a non-SQL table using this method can lead to significant information loss from the premise which can be a potential ingredient required to derive the final answer. Additionally, this is an expensive method as it requires 3X API calls as opposed to other prompting methods. Moreover, EEDP is a unified prompting strategy which integrates multiple solver elements into a single unified prompt for elicitation, extraction, decomposition and prediction.

\section{Conclusion}

In conclusion, our study delved into LLMs' mathematical reasoning in complex financial scenarios, assessing their adaptability to mixed structured tables and unstructured text. Through rigorous experimentation, we uncovered insights into their performance and limitations, presenting a tailored prompting technique that outperformed other baseline methods. Our findings advance understanding of LLMs' abilities in tackling intricate mathematical tasks within semi-structured documents, suggesting directions for future research. Please refer to appendix section \ref{sec:future_work} for future work details.

\section*{Limitations}
The scope of this work is limited by the following challenges:

\paragraph{Dataset Scarcity.} There are not many datasets dealing with numerical reasoning over semi-structured data apart from the ones from financial domain. As a future work, it would be interesting to similar analysis across various domains such as e-commerce, healthcare, sports and scientific tables from research papers, uncovering new challenges and insights. This expansion will enhance the applicability and impact of our research within the NLP community. However, creating tailored datasets for these domains presents a significant challenge.
    
For now to ensure a comprehensive evaluation of LLMs, we have integrated financial datasets that offer diverse challenges. For instance, Multihiertt evaluates model performance with intricate premise structures, providing insights into handling complex data hierarchies. ConvFinQA delves into the intricate chains of numerical reasoning within conversational question answering contexts, offering a unique perspective on dynamic data interpretation. Moreover, FinQA and TAT-QA encompass a wide array of reasoning types, with a significant portion requiring domain-specific knowledge, thereby broadening the evaluation spectrum.
    
\paragraph{Factors Isolation.} It is essential to acknowledge that there may be multiple factors influencing the performance of large language models while dealing with numerical reasoning over semi-structured data. In our analysis, we have focused on specific factors and trends, but it is important to recognize that the overall performance is affected by a multitude of variables. Marginalizing i.e. observing the trend along one while keeping the rest as constants or isolating a single factor is challenging and cannot be done with real-world data. Additionally, future investigations may benefit from simulating controlled scenarios on synthetic and counterfactual datasets to gain deeper insights into the impact of individual factors on model performance.

\paragraph{Modeling Improvement.} We emphasize our analysis on prominent models such as \texttt{GPT-4}, \texttt{GPT-3.5-TURBO}, and \texttt{PaLM 2-540B} due to their substantial size and capabilities. Notably, other open-sourced large language models like \texttt{LLaMA 2-13B}, \texttt{MAmmoTH-13B} and \texttt{Mistral-7B-Instruct} did not exhibit satisfactory performance in numerical reasoning over semi-structured data. For more detail about the the model choices refer to Appendix \ref{sec:model_selection}. This accentuates the need for exploring computationally feasible and cheaper models that can deliver remarkable performance in tasks involving numerical reasoning over heterogeneous sources of information. Future experiments with ample computational resources may involve exploring larger open-source models like OLMo, Mixtral, and DBRX, which have been recently released.

\section*{Ethics Statement}
We, the authors of this work, affirm that our work complies with the highest ethical standards in research and publication. In conducting this research, we have considered and addressed various ethical considerations to ensure the responsible and fair use of computational linguistics methodologies.  We provide detailed information to facilitate the reproducibility of our results. This includes sharing code, datasets (in our case, we deal with publicly available datasets and comply to the ethical standards mentioned by the authors of the respective works.), and other relevant resources to enable the research community to validate and build upon our work. The claims in the paper match the experimentation results, however, with \textit{black-box} large language models, a certain degree of stochasticity is expected which we attempt to minimize by keeping a fixed temperature. We describe in the fullest details the annotations, dataset splits, models used and prompting methods tried, ensuring reproducibility of our work.

\section*{Acknowledgements}
Research was sponsored by the Army Research Office and was accomplished under Grant Number W911NF-20-1-0080. The views and conclusions contained in this document are those of the authors and should not be interpreted as representing the official policies, either expressed or implied, of the Army Research Office or the U.S. Government. The U.S. Government is authorized to reproduce and distribute reprints for Government purposes notwithstanding any copyright notation herein. This work was partially funded by ONR Contract N00014-19-1-2620. Lastly, we extend our appreciation to the reviewing team for their insightful comments.

\bibliographystyle{acl_natbib}
\bibliography{acl2024}

\appendix

\section{Appendix}
\label{sec:appendix}
\subsection{How proficient are LLMs in performing simple arithmetic operations?}
\label{subsec:oom}

\begin{figure*}[!htbp]
    \centering
    \vspace{2.5em}
    \includegraphics[width= 0.95\linewidth, keepaspectratio]{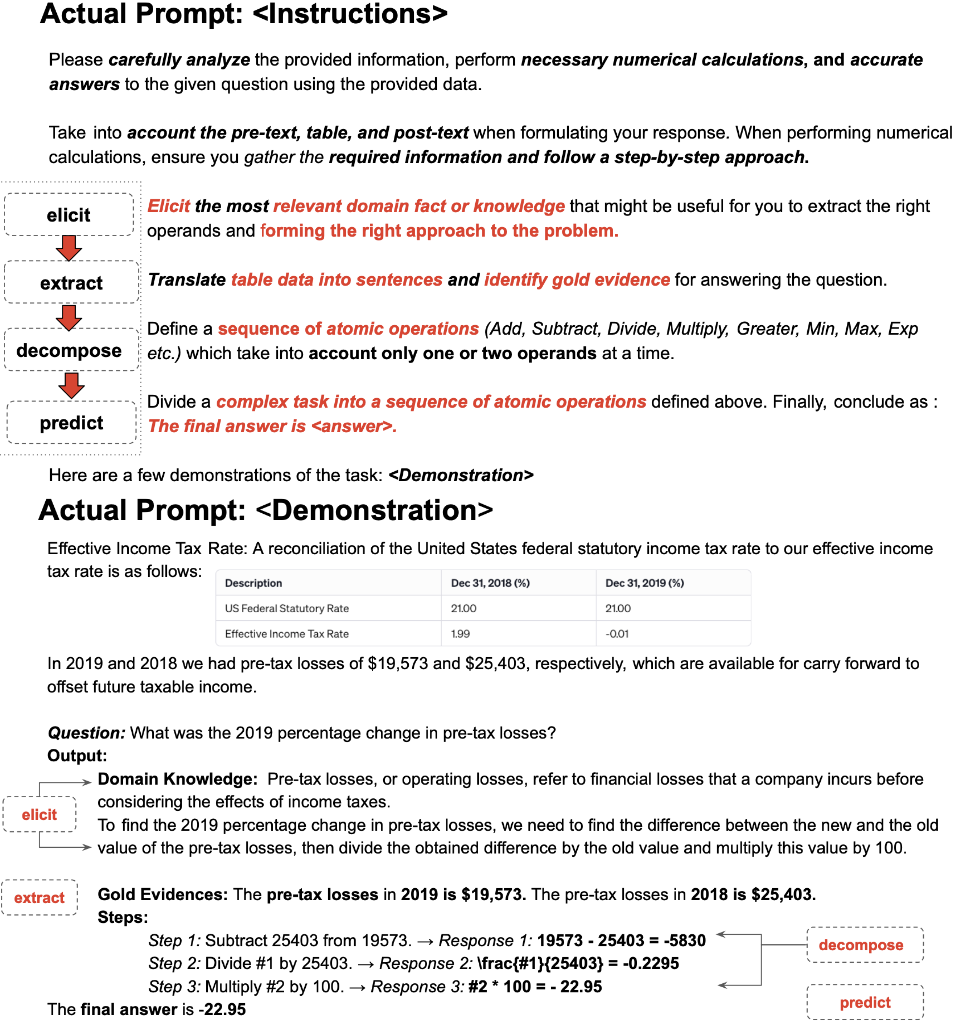}
    \vspace{-0.0em}
    \caption{Our EEDP Approach (a.) Instructions, and (b.) Demonstration.}
    \vspace{-1.0em}
    \label{fig:methodfigure}
\end{figure*}

To assess the effectiveness of Large Language Models in handling fundamental arithmetic tasks (addition (+), subtraction (-), multiplication (*), and division (/)) across operands of varying magnitudes, we generate a set of 2600 synthetic arithmetic expressions using \texttt{GPT-4}. This set includes 650 problems for each arithmetic operation. Within each operation category, we categorize tasks into groups based on a parameter denoted as \(\tau\):
\begin{center}
\vspace{-2.0em}
\begin{align*}
\tau = \text{OOM}(\;\underset{\text{\texttt{op}}}{\arg\max}\;\|\text{OOM}(\text{\texttt{op}})\|\;)
\end{align*}
\vspace{-1.5em}
\end{center}
where, $\arg\max$ selects the operand $\text{\texttt{op}}$ with the greater absolute value of the order of magnitude, and $\text{OOM}$ represents the order of magnitude. 

This approach is motivated by cognitive challenges commonly faced by humans, as they often encounter difficulties with both high and low orders of magnitude. Essentially, captures the order of magnitude of the operand with the larger absolute value among the two. For each arithmetic operation, we establish groups with \(\tau\), ranging from -6 to 6. Within each group, there are 50 arithmetic expressions. 
This systematic grouping provides a comprehensive assessment across various difficulty levels based on operand magnitudes.

\begin{figure*}[!htbp]
    \centering
    \begin{adjustbox}{max size={0.97\textwidth}{0.97\textheight}}
\includegraphics{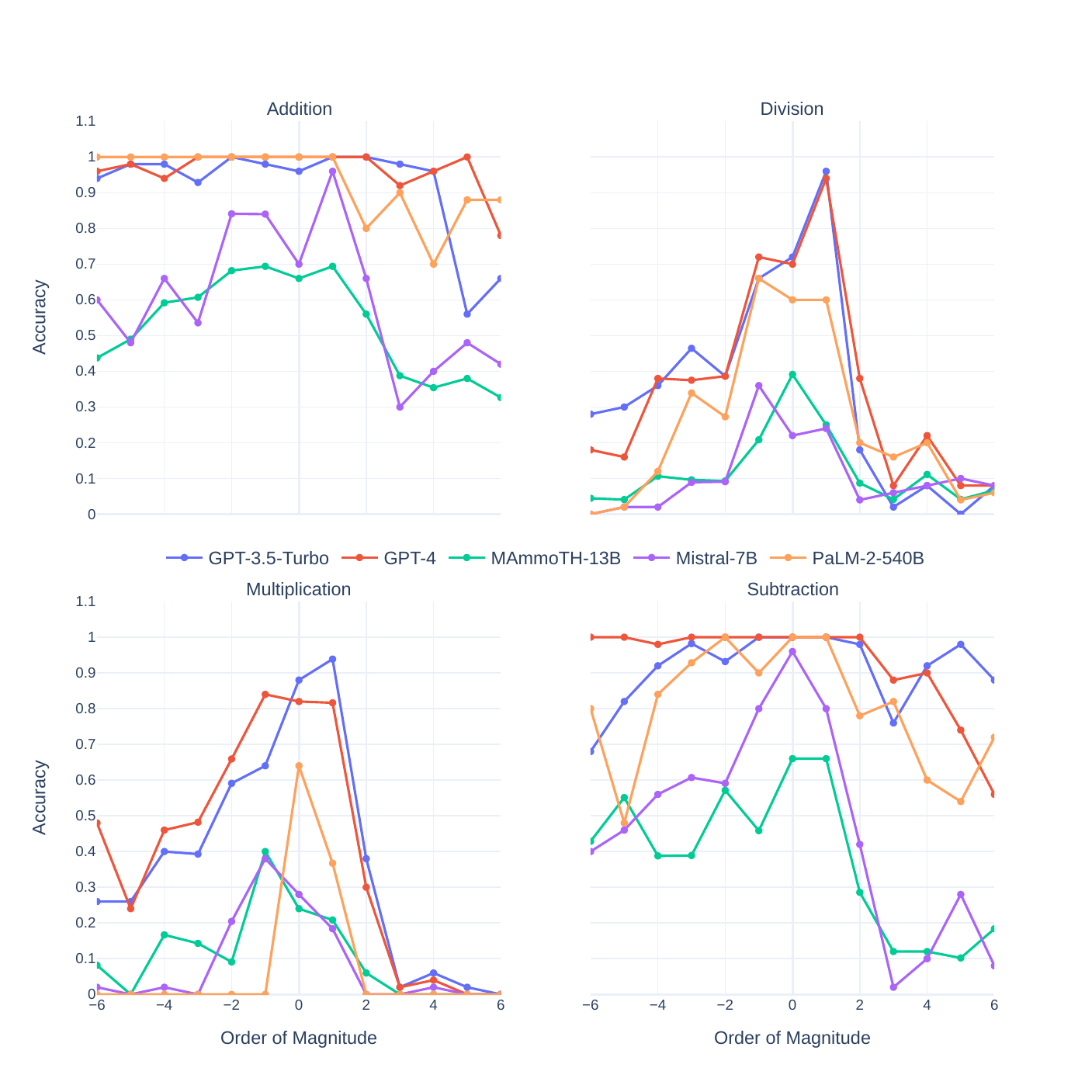}
 \end{adjustbox}
    \vspace{-1.0em}
    \caption{\small Accuracy of different arithmetic operations across different orders of magnitude.}
    \label{fig:enter-oom}
    \vspace{-2.5em}
    \label{fig:accuracy_operations}
\end{figure*}

\textit{Analysis.} Figure \ref{fig:enter-oom} illustrates that for simpler arithmetic operations like addition and subtraction, the impact of the order of magnitude of the operands is less significant compared to harder operations like multiplication and division. We observe a trend in performance accuracy with increasing and decreasing orders of magnitude. Larger models such as \texttt{GPT-4}, \texttt{GPT-3.5-TURBO}, and \texttt{PaLM 2-540B} perform significantly better on addition and subtraction tasks as compared to the multiplication and division tasks.


\begin{figure*}[!htbp]
    \centering
    \includegraphics[width= 0.97\linewidth, keepaspectratio]{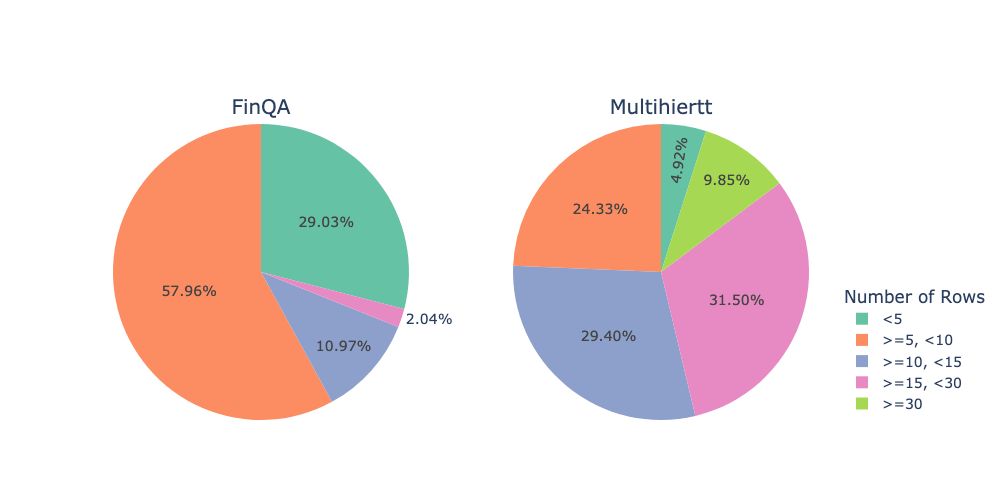}
    \vspace{-1.0em}
    \caption{ \small Sample distribution of Multihiertt \& FinQA datasets partitioned by number of rows in the supporting table.}
    \vspace{-1.0em}
    \label{fig:Number of Rows}
\end{figure*}

 \begin{figure*}[!htbp]
    \centering
    \begin{subfigure}[b]{0.5\linewidth}
        \centering
        \includegraphics[width=\linewidth, keepaspectratio]{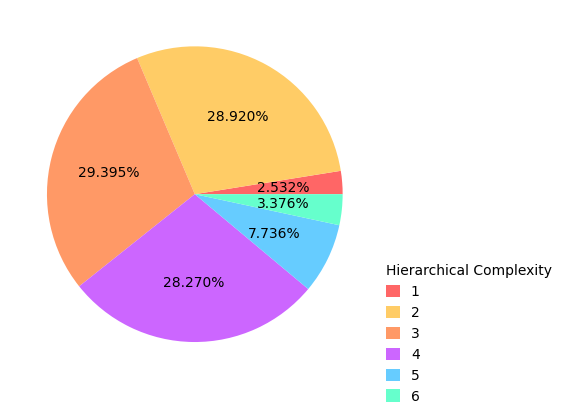}
        \caption{\small Hierarchical Complexity}
        \label{fig:hierarchical complexity}
    \end{subfigure}%
    \begin{subfigure}[b]{0.5\linewidth}
        \centering
        \includegraphics[width=\linewidth, keepaspectratio]{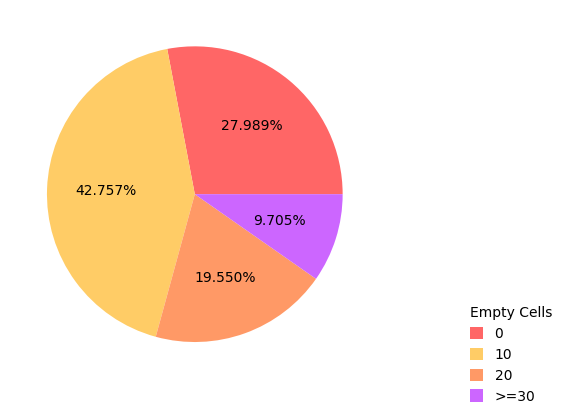}
        \caption{\small Empty Cells}
        \label{fig:empty cells}
    \end{subfigure}
    \vspace{-1.0em}
    \caption{Sample distribution of Multihiertt Dataset partitioned by (a)  hierarchical complexity of the gold evidence. (b) the percentage of empty cells in the supporting table.}
    \label{fig:hierarchical complexity}
    \vspace{-1.0em}
\end{figure*}

\subsection{Model Selection Criteria}
\label{sec:model_selection}
Our model selection process was guided primarily by resource constraints and the timeframe of our research endeavor. We aimed to identify models that represented the state-of-the-art (SOTA), such as GPT-4, or those with a specific focus on mathematical reasoning, such as \texttt{MAmmoTH}, aligning with the parameters of our project. Here's a breakdown:

\begin{enumerate}

   \item  \textbf{Resource and Budget Constraints:} Given our limited computing resources and budget, we meticulously selected models that could provide valuable insights within the confines of our project. Incorporating additional models would have been impractical given our resource limitations. 

 The number of shots, indicated by \#shots = k, was tailored to the context length of the model. Specifically, for models with a context length exceeding the input length, we standardized k to 4. For instance, we allocated 2 shots for models like LLaMA and MammoTH due to their constrained context length. However, for other models capable of accommodating larger contexts, we increased the number of shots to 4. Additionally, we used a temperature of 0 and top$_p$ = 1 for our experiments.

   \item  \textbf{Models with Mathematical Capabilities:} We prioritized models renowned for their advanced mathematical prowess, such as \texttt{MAmmoTH}, alongside state-of-the-art Large Language Models (LLMs) like GPT-4. Our goal was to gain deeper insights into the mathematical reasoning capabilities of cutting-edge models within the context of financial documents.

   \item  \textbf{Better Prompting Approaches:} Rather than focusing solely on model diversity, we concentrated on exploring a variety of prompting methods, particularly those aimed at enhancing mathematical reasoning. We believed this approach would yield more valuable insights into the performance of both LLMs and their associated prompting techniques in real-world financial tasks.

   \item  \textbf{Excluding Underperforming Models:} While we experimented with various models, such as Falcon-7B-Instruct and MPT-7B-Instruct, we found them to underperform significantly compared to models like LLaMA and Mistral. Consequently, we excluded them from detailed analysis. Future experiments with ample computational resources may involve exploring additional open-source models like OLMo, Mixtral, and DBRX, which have been recently released.

\end{enumerate}
\subsection{Other Modeling Techniques}
\label{sec:future_work}
Based on our research and the results obtained from our proposed method 'EEDP', we do have several insights that could guide future model development:
\begin{enumerate}
    \item \textbf{Domain-Specific Pre-training:} Our method "EEDP" suggests that LLMs could benefit from pre-training that focuses on extracting domain-specific knowledge. In the context of financial documents, for instance, this could involve training models on a corpus of financial texts, thereby enabling them to better understand and reason about financial concepts and terminology.
    \item \textbf{Knowledge Elicitation:}  The elicitation step in "EEDP" indicates the potential for designing LLMs that can elicit or extract relevant information from a given context more effectively. This could involve developing models that are better at identifying and focusing on key pieces of information in a document, which is crucial for accurately answering questions about the document.
    \item \textbf{Modular Modeling:} 
    Our research introduce a novel approach to the reasoning process, wherein it's broken down into modular steps. In this methodology, Large Language Models (LLMs) handle different aspects of a task in distinct stages. This division potentially enhances the overall accuracy and efficiency of the model.

For instance, the model might begin by eliciting domain-specific knowledge, then proceed to extract relevant information from the premise. Following this, it engages in reasoning about this information to answer a question and finally derives the answer, using the output from the preceding reasoning steps as a reference point.

By potentially training individual expert models to handle each specific stage, we could optimize performance for each distinct task. This modular approach allows for specialized processing of each step, thereby improving the overall performance and interpretability of the final output.

    
    
    \item \textbf{Hierarchical Structure Understanding:} Representing the input structure of the table in a better format to the LLM could be beneficial. One can also explore introducing special positional encodings, similar to those used in TAPAS, to serve as row and column IDs for each cell. This approach would differ from traditional positional encodings, which are designed to capture the inherently linear structure of textual data. TThis integration would facilitate the extraction of relvant information from the table correctly, considering its structure more effectively, avoiding information extraction errors. Another idea could be converting the premise containing the complex table and text into a common representation such as a knowledge graph. Furthermore, models specifically tuned to answer to human queries over complex documents in multiple conversational turns (like that in ConvFinQA) should also be considered, as it's a challenge for language model's to backtrack to their decisions that were made previously in the conversation.
\end{enumerate}


\subsection{Metadata Annotations Dataset Coverage}
\label{subsec:datacoverage}

\begin{figure*}[!htbp]
    \centering
    \includegraphics[width=0.85\linewidth]{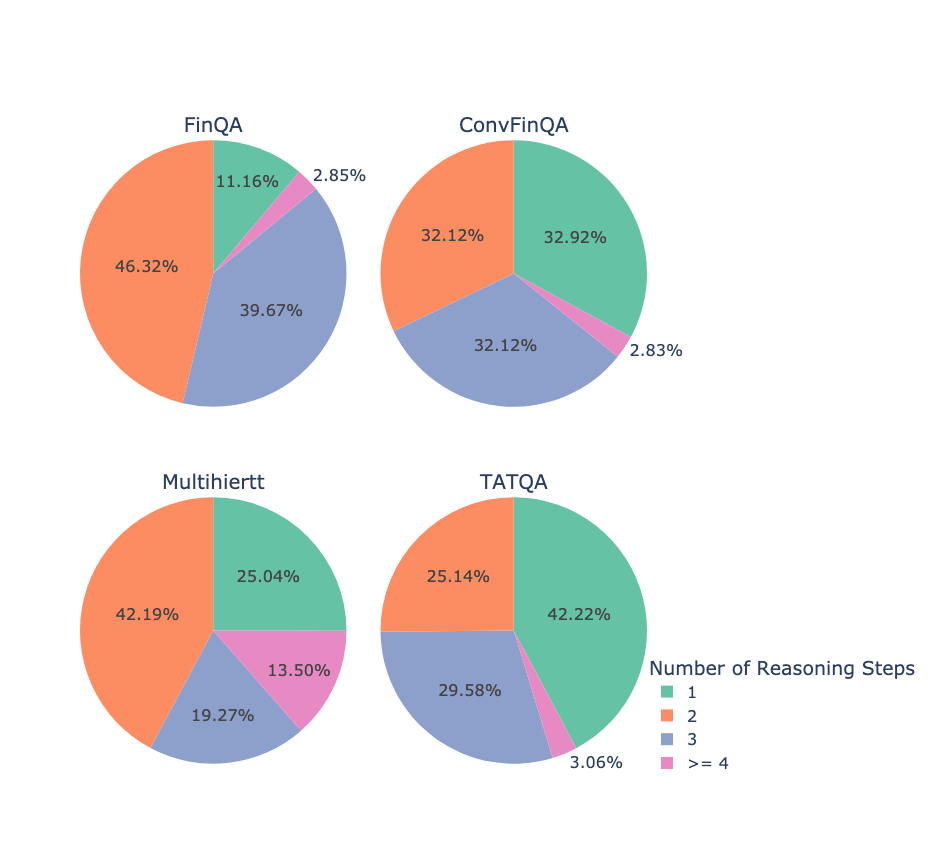}
    \vspace{-1.0em}
    \caption{ \small Sample distribution of examples in numerical reasoning on tabular datasets partitioned by the number of reasoning steps involved. Clockwise (from top-left) : FinQA, ConvFinQA, TATQA, Multihiertt.}
    \vspace{-1.5em}
    \label{fig:steps}
\end{figure*}

\begin{figure*}[!htbp]
    \centering
    \includegraphics[width=0.85\linewidth]{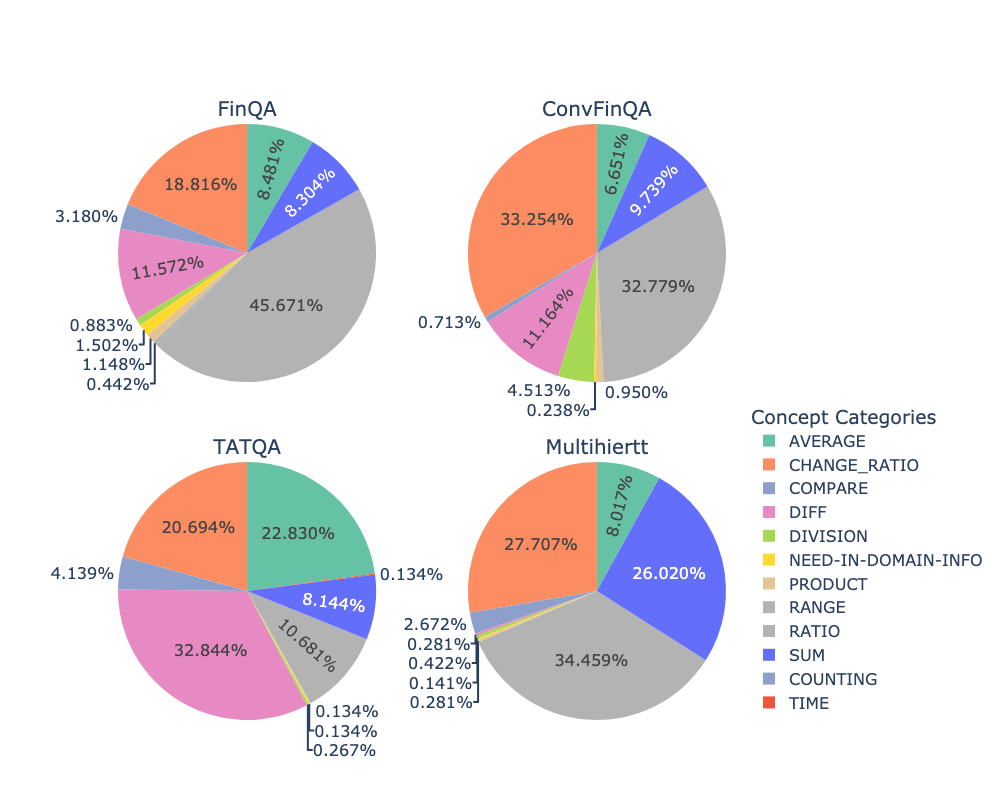}
    \vspace{-1.0em}
    \caption{ \small Sample distribution of Numerical \& Tabular Reasoning Datasets partitioned by Question Concept Category types. Clockwise (from top-left): FinQA, ConvFinQA, Multihiertt, TATQA.}
    \vspace{-1.0em}
    \label{fig:qsplit-dataset}
\end{figure*}

Figure \ref{fig:Number of Rows} displays the dataset distribution for Multihiertt and FinQA based on table length (number of rows). Figure \ref{fig:hierarchical complexity} (a) shows how we calculate hierarchical complexity for examples with multiple relevant rows at various hierarchical depths. For the distribution of missing information (empty cell proportions) across datasets, refer to Figure \ref{fig:hierarchical complexity} (b). Figure \ref{fig:steps} for the distribution of questions by reasoning steps. We define 12 mathematical concept categories, see Table \ref{table:question-categories}) and annotate each question accordingly. The dataset coverage for these categories is shown in Figure  \ref{fig:qsplit-dataset}. Our method for estimating hierarchy depth is shown in Figure \ref{fig:hierarchy-depth}. Figure \ref{eedp-prompt-template} shows a example for EEDP strategy with one shot. Figure \ref{eedp-prompt-template} shows a example for EEDP strategy with one shot.



\begin{figure*}[t]
    \centering
\begin{tcolorbox}[colback=white,arc=4mm] 
\begin{minipage}{\dimexpr\textwidth-2\fboxsep-2\fboxrule}
\centering
\medium 

\resizebox{0.65\textwidth}{!}{%
\begin{tabular}{@{}l|ccc@{}}
\toprule
\colorbox{gray!70}{\textbf{Years Ended December 31,}} & \colorbox{gray!55}{\textbf{2009}} & \colorbox{gray!55}{\textbf{2008}} \\
\midrule
\colorbox{gray!50}{\text{(in millions, except percentages)}} & & \\
\midrule
\textbf{Revenues} & & \\
\hspace{0.5cm} 
\text{Management and financial advice fees} & \$1,234 & \$1,339 \\
\hspace{0.5cm} \colorbox{gray!40}{\text{Distribution fees}} & \colorbox{gray!30}{\$1,733} & \colorbox{gray!30}{\$1,912}  \\
\hspace{0.5cm} \text{Net investment income} & \$297 & \$-43  \\
\hspace{0.5cm} \text{Other revenues} & \$85 & \$80 \\
\hspace{0.5cm} \text{Total revenues} & \$3,349 & \$3,288  \\
\midrule
\textbf{Banking and deposit interest expense} & \$133 & \$178  \\
\midrule
\textbf{Total net revenues} & \$3,216 & \$3,110  \\
\midrule
\textbf{Expenses} & &  \\
\hspace{0.5cm} \text{Distribution expenses} & \$1,968 & \$2,121 \\
\hspace{0.5cm} \text{General and administrative expense} & \$1,282 & \$1,138  \\
\hspace{0.5cm} \text{Total expenses} & \$3,250 & \$3,259  \\
\midrule
\textbf{Pretax loss} & \$-34 & \$-149  \\
\bottomrule
\end{tabular}
}
\\
\vspace{\baselineskip}
\raggedright 
\textbf{Question:} What will Distribution fees reach in 2010 if it continues to grow at its current rate? (in millions)?\\
\vspace{\baselineskip}
\textbf{Gold Evidences:} \\
\begin{itemize}
\item Table shows \colorbox{gray!40}{Distribution fees} of \colorbox{gray!70}{Years Ended December 31,} \colorbox{gray!55}{2009} \colorbox{gray!50}{(in millions, except percentages)} is \colorbox{gray!30}{\$1,733}.
\item Table shows \colorbox{gray!40}{Distribution fees} of \colorbox{gray!70}{Years Ended December 31,} \colorbox{gray!55}{2008} \colorbox{gray!50}{(in millions, except percentages)} is \colorbox{gray!30}{\$1,912}. \\
\end{itemize}
\textbf{Hierarchical Complexity:} max(4, 4) = 4
\end{minipage}
\end{tcolorbox}
    \caption{ \small This illustration demonstrates how the "hierarchical complexity" is determined for each supporting piece of evidence. The hierarchical complexity corresponds to the number of levels or tiers of information structure within the table. The top level (1) encompasses the table itself, which contains information about the results of operations in a firm's Advice \& Wealth Management segment. The second level (2) includes columns specifying the years (2009, 2008), creating a substructure within the table. The third level (3) involves the column "(in millions, except percentages)," and the fourth level (4) encompasses the rows under categories like "Revenues," "Distribution fees," "Net investment income," and so on. To locate specific values like "\$1733" and "\$1912", a retriever module needs to navigate through these four levels.}
    \vspace{-1.75em}
    
    \label{fig:hierarchy-depth}
\end{figure*}


\FloatBarrier
\twocolumn

\begin{figure*}[!htb]
\centering
\begin{tcolorbox}[colback=white,arc=4mm] 
\begin{minipage}{\dimexpr\textwidth-2\fboxsep-2\fboxrule}
\centering
\medium 

\vspace{\baselineskip}
\raggedright 
\textbf{Instruction:} Please carefully analyze the provided information, perform necessary numerical calculations, and provide accurate answers to the given question using the provided data. Take into account the pre-text, table, and post-text when formulating your response. \\
When performing numerical calculations, ensure you gather the required information and follow a step-by-step approach.\\
\begin{enumerate}
    \item Elicit the most relevant domain fact or knowledge that might be useful for you to extract the right operands and forming the right approach to the problem.
    \item Translate table data into sentences and identify gold evidence for answering the question.
    \item Define a sequence of atomic operations (Add, Subtract, Divide, Multiply, Greater, Min, Max, Exp) which take into account only two operands at a time. Divide a complex task into a sequence of atomic operations defined above. 
    \item Finally, conclude as: The final answer is \textless answer\textgreater. 
\end{enumerate}
\vspace{\baselineskip}
\raggedright 
\textbf{Demonstration:} ``Effective Income Tax Rate'': A reconciliation of the United States federal statutory income tax rate to our effective income tax rate is as follows: In 2019 and 2018 we had pre-tax losses of \$19,573 and \$25,403, respectively, which are available for carry forward to offset future taxable income. \\
\vspace{\baselineskip}
\centering
\resizebox{0.65\textwidth}{!}{%
\begin{tabular}{@{}lcc@{}}
\toprule
 & \textbf{Year Ended} & \textbf{Year Ended}  \\
\midrule
 & \textbf{December 31, 2018} & \textbf{December 31, 2019}  \\
\midrule
\textbf{United States federal statutory rate} & 21.00\% & 21.00\%  \\
\textbf{Effective income tax rate} & 1.99\% & -0.01\% \\
\bottomrule
\end{tabular}%
}

\vspace{\baselineskip}
\raggedright 
\textbf{Question:} What was the 2019 percentage change in pre-tax losses? \\
\vspace{\baselineskip}
\textbf{Response \#:} \\

\vspace{\baselineskip}

\textbf{Domain Knowledge:} Pre-tax losses, or operating losses, refer to financial losses that a company incurs before considering the effects of income taxes. To find the 2019 percentage change in pre-tax losses, we need to find the difference between the new and the old value of the pre-tax losses, then divide the obtained difference by the old value and multiply this value by 100.

\vspace{\baselineskip}

\textbf{Gold Evidences:} \\
\begin{itemize}
  \item The pre-tax losses in 2019 are \$19,573. \\
  \item The pre-tax losses in 2018 are \$25,403. \\
\end{itemize}

\vspace{\baselineskip}

\textbf{Steps:} \\
\begin{enumerate}
\item Subtract 25403 from 19573.
\begin{itemize}
\item \textbf{Response 1:} 19573 - 25403 = -5830
\end{itemize}
\item Divide \#1 by 25403.
\begin{itemize}
\item \textbf{Response 2:} $\frac{-5830}{25403} = -0.2295$
\end{itemize}
\item Multiply \#2 by 100.
\begin{itemize}
\item \textbf{Response 3:} $-0.2295 \times 100 = -22.95$
\end{itemize}

\end{enumerate}

The final answer is \textbf{-22.95\%}. \\

\end{minipage}
\end{tcolorbox}
\caption{\small A Template for our proposed prompting strategy, EEDP with 1-shot demonstration.}
\label{eedp-prompt-template}
\end{figure*}

\begin{figure*}[!htb]
\centering
\begin{tcolorbox}[colback=white,arc=4mm] 
\begin{minipage}{\dimexpr\textwidth-2\fboxsep-2\fboxrule}
\centering
\medium 
\caption*{Regulatory capital, assets, and risk-based capital ratios for JPMorgan Chase and its significant IDI subsidiaries under Basel III Standardized Transitional and Basel III Advanced Transitional at December 31, 2017, and 2016.}
\label{tab:regulatory_capital}

\resizebox{0.85\textwidth}{!}{%
\begin{tabular}{@{}lcccc@{}}
\toprule
 & \multicolumn{2}{c}{\textbf{Basel III Standardized Transitional}} & \multicolumn{2}{c}{\textbf{Basel III Advanced Transitional}} \\
\cmidrule(lr){2-3} \cmidrule(lr){4-5}
\textbf{(in millions)} & \textbf{Dec 31, 2017} & \textbf{Dec 31, 2016} & \textbf{Dec 31, 2017} & \textbf{Dec 31, 2016} \\
\midrule
Regulatory capital & & & & \\
CET1 capital & \$184,375 & \$179,319 & \$184,375 & \$179,319 \\
Tier 1 capital(a) & \$184,375 & \$179,341 & \$184,375 & \$179,341 \\
Total capital & \$195,839 & \$191,662 & \$189,419 & \$184,637 \\
Assets & & & & \\
Capital ratios(c) & & & & \\
CET1 & 13.8\% & 13.7\% & (e) & 15.0\% \\
Tier 1(a) & 13.8\% & 13.7\% & (e) & 15.0\% \\
Total & 14.7\% & 14.6\% & (e) & 15.4\% \\
Tier 1 leverage(d) & 8.7\% & 8.6\% & & 8.7\% \\
\bottomrule
\end{tabular}%
}

\vspace{\baselineskip}
\raggedright 
\textbf{Question:} What is the sum of CET1 capital, Tier 1 capital, and Total capital in 2017? (in million) \\
\vspace{\baselineskip}
\textbf{Response \#:} \\
\vspace{\baselineskip}
\textbf{Domain Knowledge:} In banking, CET1 capital, Tier 1 capital, and Total capital are used to measure a bank's financial health and stability. CET1 (Common Equity Tier 1) capital consists mainly of common shares and retained earnings. Tier 1 capital is the core capital mainly comprising shareholders' equity and disclosed reserves. Total capital includes both Tier 1 capital and Tier 2 capital, which consists of hybrid capital instruments and subordinated debt. \\
\vspace{\baselineskip}
\textbf{Gold Evidences:} \\
\begin{itemize}
\item Table shows JPMorgan Chase Bank, N.A. Basel III Standardized Transitional CET1 capital on Dec 31, 2017 as \$184,375 million.
\item Table shows JPMorgan Chase Bank, N.A. Basel III Standardized Transitional Tier 1 capital on Dec 31, 2017 as \$184,375 million.
\item Table shows JPMorgan Chase Bank, N.A. Basel III Standardized Transitional Total capital on Dec 31, 2017 as \$195,839 million.
\end{itemize}
\vspace{\baselineskip}
\textbf{Steps:} \\
\begin{enumerate}
\item Add CET1 capital and Tier 1 capital.
\begin{itemize}
\item \textbf{Response 1:} \$184,375 + \$184,375 = \$368,750
\end{itemize}
\item Add \#1 and Total capital.
\begin{itemize}
\item \textbf{Response 2:} \#1 + \$195,839 = \$564,589
\end{itemize}
\end{enumerate}
The final answer is \$564,589 million. \\
\vspace{\baselineskip}
\textbf{Ground-Truth Steps:} 
\$184,375 + \$184,375 + \$195,839 + \$184,375 + \$184,375 + \$189,419 = \$1,122,758 \\
\vspace{\baselineskip}
\textbf{Reason:} The model missed adding JPMorgan Chase Bank, N.A. Basel III Advanced Transitional CET1 capital (\$184,375), Tier 1 (\$184,375), and Total Capital (\$189,419) in 2017.
\end{minipage}
\end{tcolorbox}
\caption{Error due to missing evidences}
\label{Missing Evidences}
\end{figure*}

\begin{figure*}[!htb]
\centering
\begin{tcolorbox}[colback=white,arc=4mm] 
\begin{minipage}{\dimexpr\textwidth-2\fboxsep-2\fboxrule}
\centering
\medium 
\resizebox{0.85\textwidth}{!}{%
\begin{tabular}{@{}lcccccc@{}}
\toprule
\textbf{Year} & \textbf{Life/Fin.} & \textbf{Gen.} & \textbf{Rtd.} & \textbf{Fin.} & \textbf{Asset} & \textbf{Total} \\
 & \textbf{Ins.} & \textbf{Ins.} & \textbf{Services} & \textbf{Services} & \textbf{Mngmt} &  \\
 \midrule
2006 &  &  &  &  &  &  \\
\textbf{Fixed Maturities:} &  &  &  &  &  &  \\
Bonds available for sale & \$67,994 & \$288,018 & \$1,357 & \$29,500 & — & \$386,869 \\
Bonds held to maturity & \$21,437 & — & — & — & — & \$21,437 \\
Bond trading securities & \$1 & \$10,835 & — & — & — & \$10,836 \\
\textbf{Equity Securities:} &  &  &  &  &  &  \\
Common stocks & \$4,245 & \$8,705 & — & \$226 & \$80 & \$13,256 \\
Preferred stocks & \$1,884 & \$650 & \$5 & — & — & \$2,539 \\
\textbf{Other Assets:} &  &  &  &  &  &  \\
Mortgage and other loans & \$17 & \$21,043 & \$2,398 & \$4,884 & \$76 & \$28,418 \\
Flight equipment & — & — & \$39,875 & — & — & \$39,875 \\
\textbf{Total Invested Assets} & \$116,059 & \$428,825 & \$185,404 & \$68,280 & \$3,373 & \$801,941 \\
\bottomrule
\end{tabular}%
}

\vspace{\baselineskip}
\raggedright 
\textbf{Question:} What is the proportion of Asset Management to the total Mortgage and other loans receivable, net of allowance in 2006? \\
\vspace{\baselineskip}
\textbf{Response \#:} \\

\vspace{\baselineskip}

\textbf{Domain Knowledge:} Proportion refers to the relationship between two quantities, where one quantity is considered in relation to the other. To find the proportion of Asset Management to the total Mortgage and other loans receivable, net of allowance in 2006, we need to divide the value of Asset Management by the total value of Mortgage and other loans receivable, net of allowance in 2006. \\

\vspace{\baselineskip}

\textbf{Gold Evidences:} \\
\begin{itemize}
\item Table shows Asset Management of Total invested assets in 2006 is \$68,280.
\item Table shows Mortgage and other loans receivable, net of allowance (in millions) in 2006 is \$28,418.
\end{itemize}

\vspace{\baselineskip}

\textbf{Steps:} \\
\begin{enumerate}
\item Divide Asset Management by the total Mortgage and other loans receivable, net of allowance in 2006.
\begin{itemize}
\item \textbf{Response 1:} $\frac{68280}{28418} = 2.4024$
\end{itemize}
\end{enumerate}
The final answer is 2.4024. \\

\vspace{\baselineskip}
\textbf{Ground-truth steps:}
$\frac{4884.0}{28418.0} = 0.172$ \\
\vspace{\baselineskip}
\textbf{Reason:} Asset Management in 2006 is taken as \$68,280.0 by the model instead of \$4,884.0 due to error in the extraction of the ground-truth.
\end{minipage}
\end{tcolorbox}
\caption{Error due to wrong evidences}
\label{Wrong Evidences}
\end{figure*}

\begin{figure*}[!htb]
\centering
\begin{tcolorbox}[colback=white,arc=4mm] 
\begin{minipage}{\dimexpr\textwidth-2\fboxsep-2\fboxrule}
\centering
\medium 

\resizebox{0.88 \textwidth}{!}{%
\begin{tabular}{@{}lccccc@{}}
\toprule
 & \textbf{2013} & \textbf{2014} & \textbf{2015} & \textbf{2016} & \textbf{2017} \\
 \midrule
\textbf{Masco} & \$138.48 & \$155.26 & \$200.79 & \$227.08 & \$318.46 \\
\textbf{S\&P 500 Index} & \$132.04 & \$149.89 & \$151.94 & \$169.82 & \$206.49 \\
\textbf{S\&P Industrials Index} & \$140.18 & \$153.73 & \$149.83 & \$177.65 & \$214.55 \\
\textbf{S\&P Consumer Durables \& Apparel Index} & \$135.84 & \$148.31 & \$147.23 & \$138.82 & \$164.39 \\
\bottomrule
\end{tabular}%
}

\vspace{\baselineskip}
\raggedright 
\textbf{Question:} What was the difference in percentage cumulative total shareholder return on Masco common stock versus the S\&P 500 Index for the five-year period ended 2017? \\
\vspace{\baselineskip}
\textbf{Response \#:} \\

\vspace{\baselineskip}

\textbf{Domain Knowledge:} Cumulative total shareholder return is a measure of the overall performance of a stock over a given period of time. It includes the change in the stock's price as well as any dividends paid to shareholders during that time. \\

\vspace{\baselineskip}

\textbf{Gold Evidences:} \\
\begin{itemize}
\item The cumulative total shareholder return for Masco in 2017 is \$318.46.
\item The cumulative total shareholder return for the S\&P 500 Index in 2017 is \$206.49.
\end{itemize}

\vspace{\baselineskip}

\textbf{Steps:} \\
\begin{enumerate}
\item Divide the cumulative total shareholder return for Masco in 2017 (318.46) by the cumulative total shareholder return for the S\&P 500 index in 2017 (206.49)
\begin{itemize}
\item \textbf{Response 1:} 318.46 / 206.49 = 1.5422
\end{itemize}
\item Subtract 1 from the quotient obtained in \#1
\begin{itemize}
\item \textbf{Response 2:} 1.5422 - 1 = 0.5422
\end{itemize}
\item Multiply the value obtained in \#2 with 100 to find the percentage difference
\begin{itemize}
\item \textbf{Response 3:} 0.5422 * 100 = 54.22
\end{itemize}

\end{enumerate}
The final answer is 54.22\%. \\

\vspace{\baselineskip}
\textbf{Ground-truth steps:} [((318.46 - 100.0)/100.0)*100.0 - ((206.49 - 100.0)/100.0)*100.0] = 111.97 \\
\vspace{\baselineskip}
\textbf{Reason:} Wrong formula for return on investment
\end{minipage}
\end{tcolorbox}
\caption{Example of an error caused due to insufficient domain knowledge}
\label{DK-Deficit}
\end{figure*}

\begin{figure*}[!htb]
\centering
\begin{tcolorbox}[colback=white,arc=4mm] 
\begin{minipage}{\dimexpr\textwidth-2\fboxsep-2\fboxrule}
\centering
\medium 
\resizebox{0.75\textwidth}{!}{%
\begin{tabular}{@{}lccc@{}}
\toprule
  & \textbf{Accr. Yield} & \textbf{Carrying Amt.} & \textbf{Allowance} \\
 & \textbf{(Beginning)} & \textbf{(Beginning)} & \textbf{(Beginning)} \\
 \midrule 
\textbf{Beginning Balance} & \$27 & \$920 & \$95 \\
\textbf{Purchases} & \$1 & \$130 & \$0 \\
\textbf{Disposals/Payments Received} & \$-11 & \$-594 & \$0 \\
\textbf{Accretion} & \$-44 & \$44 & \$0 \\
\textbf{Builds (Reductions) to the Allowance} & \$0 & \$0 & \$-18 \\
\textbf{Increase to Expected Cash Flows} & \$-2 & \$19 & \$0 \\
\textbf{FX/Other} & \$17 & \$-50 & \$0 \\
\textbf{Balance at December 31, 2010} & \$116 & \$469 & \$77 \\
\bottomrule
\end{tabular}%
}

\vspace{\baselineskip}
\raggedright 
\textbf{Question:} What percentage did this total represent in relation to the carrying amount? \\
\vspace{\baselineskip}
\textbf{Response \#:} \\

\vspace{\baselineskip}

\textbf{Domain Knowledge:} To find the percentage of a value (A) in relation to another value (B), divide the first value (A) by the second value (B) and multiply it by 100. To find the combined total, add the two values. \\

\vspace{\baselineskip}

\textbf{Gold Evidences:} \\
\begin{itemize}
\item The carrying amount of the company's purchased distressed loan portfolio at December 31, 2010, was \$392 million.
\item The net allowance was \$77 million as of December 31, 2010. \\

\end{itemize}

\vspace{\baselineskip}

\textbf{Steps:} \\
\begin{enumerate}
\item Find the combined total of net allowance and the carrying amount.
\begin{itemize}
\item \textbf{Response 1:} 392 + 77 = 469
\end{itemize}
\item Calculate the percentage of net allowance in relation to the carrying amount.
\begin{itemize}
\item \textbf{Response 2:} $\frac{77}{392} \times 100 = 19.64$
\end{itemize}
\item Calculate the percentage of the combined total in relation to the carrying amount.
\begin{itemize}
\item \textbf{Response 3:} $\left( \frac{469}{392} \right) \times 100 = 119.64$
\end{itemize}

\end{enumerate}

The final answer is 119.64\%. \\

\vspace{\baselineskip}
\textbf{Ground-truth steps:} [(77/(392 + 77))*100] = 16.42\%  
\\
\vspace{\baselineskip}
\textbf{Reason:} By the phrase in relation to the question expected a ratio of $quantity_1/quantity_2$ but the model returned $quantity_2/quantity_1$
\end{minipage}
\end{tcolorbox}
\caption{Example of an error caused due to question misinterpretation}
\label{Incorrect-Interpret}
\end{figure*}

\begin{figure*}[!htb]
\centering
\begin{tcolorbox}[colback=white,arc=4mm] 
\begin{minipage}{\dimexpr\textwidth-2\fboxsep-2\fboxrule}
\centering
\medium 
\resizebox{0.65\textwidth}{!}{%
\begin{tabular}{@{}lcc@{}}
\toprule
 & \textbf{2001} & \textbf{2000} \\
 \midrule
\textbf{Gain on Sales of Depreciable Properties} & \$45428 & \$52067 \\
\textbf{Gain on Land Sales} & \$5080 & \$9165 \\
\textbf{Impairment Adjustment} & \$-4800 (4800) & \$-540 (540) \\
\textbf{Total} & \$45708 & \$60692 \\
\bottomrule
\end{tabular}%
}

\vspace{\baselineskip}
\raggedright 
\textbf{Question:} What is the percent change in general and administrative expense from 2000 to 2001? \\
\vspace{\baselineskip}
\textbf{Response \#:} \\

\vspace{\baselineskip}

\textbf{Domain Knowledge:} To find the percent change between two values, we need to find the difference between the new and old values, divide the difference by the old value, and multiply the result by 100. \\

\vspace{\baselineskip}

\textbf{Gold Evidences:} \\
\begin{itemize}
\item General and administrative expense decreased from \$21.1 million in 2000 to \$15.6 million for the year ended December 31, 2001. \\

\end{itemize}

\vspace{\baselineskip}

\textbf{Steps:} \\
\begin{enumerate}
\item Calculate the change in general and administrative expense by subtracting the 2001 value (\$15.6 million) from the 2000 value (\$21.1 million).
\begin{itemize}
\item \textbf{Response 1:} 21.1 - 15.6 = 5.5
\end{itemize}
\item Divide the difference obtained in Step 1 by the general and administrative expense in 2000 (\$21.1 million).
\begin{itemize}
\item \textbf{Response 2:} $\frac{5.5}{21.1} = 0.2607$
\end{itemize}
\item Multiply the result obtained in Step 2 by 100 to get the percent change.
\begin{itemize}
\item \textbf{Response 3:} $0.2607 \times 100 = 26.07$
\end{itemize}

\end{enumerate}

The final answer is 26.07\%. \\

\vspace{\baselineskip}
\textbf{Ground-truth steps:} [((15.6 - 21.1)/21.31)*100] = -26.07\% 
\\
\vspace{\baselineskip}
\textbf{Reason:} Incorrect Instantiation of variables resulting in a value with a flipped sign
\end{minipage}
\end{tcolorbox}
\caption{Example of an error caused due to incorrect instantiation of variables}
\label{Incorrect-Instantiation}
\end{figure*}

\begin{figure*}[!htb]
\centering
\begin{tcolorbox}[colback=white,arc=4mm] 
\begin{minipage}{\dimexpr\textwidth-2\fboxsep-2\fboxrule}
\centering
\medium 

\resizebox{0.65\textwidth}{!}{%
\begin{tabular}{@{}lccccc@{}}
\toprule
 & \textbf{2013} & \textbf{2014} & \textbf{2015} & \textbf{2016} & \textbf{2017} \\
\midrule
\textbf{CME Group Inc.} & \$164.01 & \$194.06 & \$208.95 & \$279.85 & \$370.32 \\
\textbf{S\&P 500} & \$132.39 & \$150.51 & \$152.59 & \$170.84 & \$208.14 \\
\textbf{Peer Group} & \$176.61 & \$187.48 & \$219.99 & \$249.31 & \$323.23 \\
\bottomrule
\end{tabular}%
}

\vspace{\baselineskip}
\raggedright 
\textbf{Question:} What is the annualized return for s\&p 500 from 2012 to 2017? \\
\vspace{\baselineskip}
\textbf{Response \#:} \\

\vspace{\baselineskip}

\textbf{Domain Knowledge:} The annualized return refers to the average rate of return over a given period for an investment, which has been adjusted for compounding. To find the annualized return for the S\&P 500 from 2012 to 2017, we need to use the formula:

\[ \text{Annualized return} = \left( \left( \frac{\text{Ending value}}{\text{Starting value}} \right)^{\frac{1}{\text{Number of years}}} - 1 \right) \] \\

\vspace{\baselineskip}

\textbf{Gold Evidences:} \\
\begin{itemize}
\item Investment was \$100 on December 31, 2012;Value of S\&P 500 investment on December 31, 2017 is \$208.14. \\

\end{itemize}

\vspace{\baselineskip}

\textbf{Steps:} \\
\begin{enumerate}
\item Divide the ending value (208.14) by the starting value (100)
\begin{itemize}
\item \textbf{Response 1:} $\frac{208.14}{100} = 2.0814$
\end{itemize}
\item Calculate the number of years elapsed (2017 - 2012)
\begin{itemize}
\item \textbf{Response 2:} 2017 - 2012 = 5
\end{itemize}
\item Raise the value obtained in \#1 to the power of (1/number of years)
\begin{itemize}
\item \textbf{Response 3:} $(2.0814) ^ {\frac{1}{5}} = 1.1606$
\end{itemize}
\item Subtract 1 from the value obtained in \#3
\begin{itemize}
\item \textbf{Response 4:} 1.1606 - 1 = 0.1606
\end{itemize}
\item Multiply the value obtained in \#4 by 100 to get the annualized return in percentage
\begin{itemize}
\item \textbf{Response 5:} 0.1606 * 100 = 16.06
\end{itemize}

\end{enumerate}

The final answer is 16.06\%. \\

\vspace{\baselineskip}
\textbf{Ground-truth steps:} [$(208.14/100)^(1/(2017-2012))$ - 1]  = 0.1577 * 100 = 15.77
\\
\vspace{\baselineskip}
\textbf{Reason:} Inaccurate calculation
\end{minipage}
\end{tcolorbox}
\caption{Example of an error caused due to imprecise calculations}
\label{Incorrect-Calculation}
\end{figure*}

\end{document}